\documentclass[pdflatex,sn-mathphys-num]{sn-jnl}

\usepackage{graphicx}%
\usepackage{multirow}%
\usepackage{amsmath,amssymb,amsfonts}%
\usepackage{amsthm}%
\usepackage{mathrsfs}%
\usepackage[title]{appendix}%
\usepackage{xcolor}%
\usepackage{textcomp}%
\usepackage{manyfoot}%
\usepackage{booktabs}%
\usepackage{algorithm}%
\usepackage{algorithmicx}%
\usepackage{algpseudocode}%
\usepackage{listings}%

\theoremstyle{thmstyleone}%
\theoremstyle{thmstyletwo}%

\theoremstyle{thmstylethree}%

\begin{document}

\title[A Learning-Based Ansatz Satisfying Boundary Conditions in Variational Problems]{A Learning-Based Ansatz Satisfying Boundary Conditions in Variational Problems}


\author*[1]{\fnm{Rafael} \sur{Florencio}}\email{rfdiaz@ujaen.es}
\equalcont{These authors contributed equally to this work.}

\author[2]{\fnm{Julio} \sur{Guerrero}}\email{jguerrer@ujaen.es}
\equalcont{These authors contributed equally to this work.}

\affil*[1]{\orgdiv{Departamento de Matem\'aticas}, \orgname{Universidad de Ja\'en}, \orgaddress{\street{Edificio de Servicios Generales - Campus Cient\'ifico Tecnol\'ogico de Linares. Avda. de la Universidad (Cintur\'on Sur), s/n}, \city{Linares}, \postcode{23700}, \state{Ja\'en}, \country{Espa\~{n}a}}}

\affil[2]{\orgdiv{Departamento de Matem\'aticas}, \orgname{Universidad de Ja\'en}, \orgaddress{\street{Campus Las Lagunillas}, \city{Ja\'en}, \postcode{23071}, \state{Ja\'en}, \country{Espa\~{n}a}}}


\abstract{\hspace{2em} Recently, innovative adaptations of the Ritz method incorporating deep learning have been developed, known as the Deep Ritz Method. This approach employs a neural network as the trial function for variational problems. However, the neural network does not inherently satisfy the boundary conditions of the variational problem. To address this issue, the Deep Ritz Method introduces a penalty term into the functional, which is strongly dependent on hyperparameters and may lead to misleading results during the optimization process.
	
\hspace{2em} In this work, we propose an ansatz that inherently satisfies the boundary conditions of the variational problem, thereby eliminating the need for penalty terms. A key contribution of this study is that all supporting theorems and corollaries are established in Sobolev norms, which constitute the natural framework for variational problems, as the functional depends explicitly on the solution and its derivatives. This provides a rigorous justification for the expressiveness and admissibility of the proposed ansatz within the Ritz method.
	
The results demonstrate that the proposed ansatz not only avoids misleading optimization outcomes but also reduces complexity while maintaining accuracy, highlighting its practical effectiveness for solving variational problems.}




\maketitle

\section{Introduction}\label{sec1}
\hspace{2em} Partial differential equations (PDEs) arise in many scientific branches. In some cases problems, the PDE can be derived from a functional associated to a variational problem. In these cases, the solutions of the PDE problem can be recognized as critical points of the functional \cite{Evansbook}. This forms the foundation of the calculus of variations \cite{Charles}-\cite{Rindler}, and its solutions involve the well-known Euler-Lagrange equations, which allow for a transition to the PDE approach. This approach is frequently found in physics \cite{Moore}. In fact, there are many versions of this approach depending on the different fields of physics (classical mechanics \cite{Goldstein},\cite{Marsden}; quantum mechanics \cite{Feynman},\cite{Kleinert}; particle physics \cite{FeynmanHibbs},\cite{Schwartz}; general relativity \cite{Foster},\cite{Carroll}, etc.). These versions of the approach are known as Action Principles and constitute the core of the different fields of physics. An Action Principle starts with a scalar function called a Lagrangian, which describes the physical system. The accumulated value of this Lagrangian function between two states of the system is called the action. The Action Principles assume that, among all possibilities, the actual evolution of the physical system produces a critical point of the action.

When exact solutions are not possible, approximation techniques are usually employed, such as the Finite Element Method (FEM) \cite{strangbook}, \cite{Ern}, Galerkin Method \cite{Grekas}, and Ritz Method \cite{Bailey}, \cite{Patnaik}. The Ritz Method is an interesting approximation technique which consists of choosing a finite number of admissible trial functions such that the solution $y(x)$ can be determined as a linear combination of these finite trial functions. Recently, novel versions of the Ritz Method integrated with deep learning have been explored. This is known as the Deep Ritz Method \cite{DRM}-\cite{Liu}. This method is based on the Universal Approximation Theorem, which essentially states that a neural network with a hidden layer is dense in subsets of a Sobolev space \cite{univapp}-\cite{univapp3}. However, the boundary conditions of the variational problem are not satisfied by the neural network. To address this non-fulfillment of the boundary conditions, the Deep Ritz Method redefines the action by adding a penalty term.

On one hand, the added penalty term can cause the action to become a non-convex function. This is a drawback, as finding local minima and saddle points for the action is a non-trivial task. On the other hand, this penalty term is weighted by a hyperparameter that must be adjusted to assign more or less importance to the boundary conditions in the optimization process. This adjustment makes the convergence rate strongly dependent on this hyperparameter.

In recent works \cite{Berr23,Sha25}, several ansatzes have been proposed to enforce the boundary conditions of admissible functions in PINNs (Physics-Informed Neural Networks) and VPINNs (Variational Physics-Informed Neural Networks), thereby avoiding the need for penalty terms. However, these works only provide numerical experiments and do not include theoretical results supporting the proposed ansatzes. In this work, we revisit the idea of using ansatzes that exactly satisfy the boundary conditions, but we introduce a simpler construction together with the corresponding theoretical results that justify the proposed approach in Sobolev spaces. This ansatz consists of one term that enforces the boundary behaviour and another term given by the product of a neural network and a function $g$, chosen so that their ratio remains differentiable inside the domain. The function $g$ can be adjusted to vanish at the boundary of the domain, ensuring that the boundary conditions are satisfied. For the rest of the domain, the behavior of the ansatz is dominated by the product of the neural network and the function $g$. It is important to point out that, since the Lagrangian usually depends of the admisible functions and their derivatives, Sobolev spaces are the natural framework to demonstrate the theorems that support the proposed ansatzes.  

Throughout this work, we demonstrate that the product of a polynomial and a single-hidden-layer neural network is dense in the relevant Sobolev spaces, provided that the polynomial is chosen so that their ratio remains differentiable inside the domain. This property allows us to consider the proposed ansatz as an admissible function to approximate the solution of the variational problem using the Ritz method approach. It is important to point out that the conditions ensuring the validity of using this type of ansatz had not been theoretically established before. Moreover, we present numerical experiments showing that the proposed ansatz significantly reduces the complexity of the neural network and prevents misleading outcomes in the optimization of the action functional (for instance, obtaining lower action values at the cost of not satisfying the boundary conditions exactly).

The remainder of the paper is organized as follows: The problem definition, theorems, and lemmas required are presented in Section 2. The ansatz and proposed method are described in Section 3. Several examples and comparative results are shown to compare the performance of the proposed method with the well-established Deep Ritz Method. Finally, concluding remarks are presented in Section 7.

\section{Theoretical Foundations: Universal Approximation with Polynomial Factors}\label{sec2}

\hspace{2em} Formally, a Lagrangian function is defined as follows: 

\begin{align}
	\label{eq1}
	\mathcal{L}:\left[\mathcal{C}^1(\Omega)\right]\times\mathbb{R} \rightarrow\mathbb{R}
\end{align}

Where we assume that $\Omega\subset\mathbb{R}^n$ is compact. The action is defined as the accumulated value of the Lagrangian over a domain $\Omega$:

\begin{align}
	\label{eq2}
	S=\int_{\Omega}{\mathcal{L}\left(y(x),\frac{\partial y(x)}{\partial x},x\right)dx}
\end{align}
 
\noindent where $y(x)\in \mathcal{C}^1(\Omega)$ belongs to the set of admissible functions such that $y(x)=b(x)$ $\forall x\in\partial\Omega\subset\mathbb{R}^{n}$ and $\left(\frac{\partial y(x)}{\partial x}\right)_{j}=\frac{\partial y}{\partial x_j}$ with $j=1,\ldots,n$. The action principles assume that the real evolution $y(x)$ produces a stationary critical point of the action $S$ (i.e., $\delta S=0$, where $\delta S$ denotes the functional derivative). This constitutes the class of variational problems, and their solutions involve the well-known Euler-Lagrange equations, which allow for a transition to the PDE approach \cite{Evansbook}.

When exact solutions are not possible, approximation techniques are usually employed. One interesting approximation technique is the Ritz Method \cite{strangbook}. Recently, a novel version of the Ritz Method integrated with deep learning has been explored, known as the Deep Ritz Method \cite{DRM}. This method is based on the Universal Approximation Theorem \cite{univapp} and its extension \cite{univapp2}. In \cite{univapp}, it is established that a single-hidden-layer neural network is dense in the space $\mathcal{C}(\Omega)$ of continuous functions. The work \cite{univapp2} provides a generalization, showing that single-hidden-layer neural networks with differentiable activation functions are dense in $\mathcal{C}^{m}(\Omega)$ for compact domains $\Omega\subset\mathbb{R}^{n}$. Consequently, this extended version of the theorem can also be formulated in terms of the Sobolev norms \cite{Adams,Brezis}:\\ 

\textbf{Definition 1.}
	\emph{
		Let $m,\mathrm{p}\in\mathbb{N}$ and $1\le p<\infty$.
		Let $f\in\mathcal{C}^m(\Omega)$. 
		Then, Sobolev norms are defined as:}
	\begin{align}
		\label{eq3}
		\|f\|_{m,\mathrm{p},\Omega}
		=
		\left(
		\sum_{|\alpha|\le m}
		\|D^\alpha f\|_{L^\mathrm{p}(\Omega)}^{\mathrm{p}}
		\right)^{1/\mathrm{p}}
	\end{align}
	\begin{align}
		&\label{eq3b}
		\|f\|_{m,\infty,\Omega}
		=
		\max_{|\alpha|\le m}
		\|D^\alpha f\|_{L^\infty(\Omega)}
	\end{align}
	\emph{where $D^\alpha$ denotes multi-index notation for partial derivatives and $||\cdot||_{L^\mathrm{p}(\Omega)}$ is $L^\mathrm{p}$-norm in $\Omega$}.\\	
	 	
According to \cite{univapp2} the universal approximation theorem can be stated by using Sobolev norms as follows:\\

\textbf{Universal Approximation Theorem for Sobolev Spaces \cite{univapp2}.}
\emph{
	Let $m\in\mathbb{N}$ and $1\le p<\infty$.
	Let $\sigma\in\mathcal{C}^m(\mathbb{R})$ be a non-constant and bounded function, 
	and let $f\in \mathcal{C}^m(\Omega)$, where $\Omega\subset\mathbb{R}^n$ is compact.
	Then, for any $\varepsilon>0$, there exist $N\in\mathbb{N}$, 
	$w_l\in\mathbb{R}^n$, $b_l\in\mathbb{R}$, $\alpha_l\in\mathbb{R}$, $l=1,\ldots,N$, such that}
\begin{align}
	\label{eq4}
	\bigg\|f(x)-\sum_{l=1}^{N}\alpha_l \sigma(w_l^T x +b_l)\bigg\|_{m,p,\Omega}<\varepsilon,
\end{align}
\emph{where $A^T$ denotes the transpose of $A$.}\\

According to \cite{univapp2}, the Universal Approximation Theorem also holds in Sobolev spaces, where the approximation error is measured with the norm $\|\cdot\|_{m,\mathrm{p},\Omega}$ instead of the supremum norm $\|\cdot\|_{\infty}$ as it is stated in \cite{univapp}. This point is important because, for variational problems with Dirichlet boundary conditions, the norm $\|\cdot\|_{m,\mathrm{p},\Omega}$ is the natural norm to measure approximation errors. In particular, since the Lagrangian function depends of $y(x)$ and $\frac{\partial y(x)}{\partial x}$, the approximation $y(x)\approx \sum_{l=1}^{N}{\alpha_l \sigma(w_l^T x +b_l)}$ made in the Deep Ritz Method should be measured by the norm $\|y(x)-\sum_{l=1}^{N}{\alpha_l \sigma(w_l^T x +b_l)}\|_{1,\mathrm{p},\Omega}$ for a fixed value of $N$, and searches for $w_l\in\mathbb{R}^n$, $b_l\in \mathbb{R}$, and $\alpha_l\in\mathbb{R}$ such that $\delta S=0$ should be done using a suitable optimization algorithm.\\
\indent Note that the boundary condition $y(x)=b(x)$ $\forall x\in\partial\Omega\subset\mathbb{R}^{n}$ is not guaranteed, since the parameters of the neural network $w_l\in\mathbb{R}^n$, $b_l\in \mathbb{R}$, and $\alpha_l\in\mathbb{R}$ are only optimized to ensure $\delta S=0$. To address the absence of the boundary condition, the Deep Ritz Method redefines the action using a penalty method as
\begin{align}
	\label{eq5}
	S_{\mathrm{DR}}=\int_{\Omega}{\mathcal{L}\left(y(x),\frac{\partial y(x)}{\partial x},x\right)dx}+\beta\int_{\partial\Omega}{(y(x)-b(x))^2dx}
\end{align}
\noindent where $\beta$ is an adjustable parameter that weights the importance of the boundary condition in the optimization process to achieve $\delta S_{\mathrm{DR}}=0$. The main drawback of the penalty method is that the boundary condition is not exactly satisfied but only approximated.

In order to ensure the exact satisfaction of the boundary conditions, we propose an ansatz that includes the product of a single-hidden-layer neural network and a suitably chosen function. With this construction, we present the following theorem, which extends the Universal Approximation Theorem to products of functions and single-hidden-layer neural networks:\\

\textbf{Theorem 1.} \emph{Let $m\in\mathbb{N}$ and $1\le p<\infty$. Let $\sigma\in\mathcal{C}^m(\mathbb{R})$ be a non-constant and bounded function and $f,g\in\mathcal{C}^m(\Omega)$, such that $f/g\in\mathcal{C}^m(\Omega)$ where $\Omega\subset\mathbb{R}^n$ is compact. Then, for any $\varepsilon'>0$, there exist $N\in\mathbb{N}$, $w_l\in\mathbb{R}^n$, $b_l\in \mathbb{R}$, and $\alpha_l\in\mathbb{R}$ with $l=1,\ldots,N$ such that}
\begin{align}
	\label{eq6}
	&\bigg\|f(x)-g(x)\sum_{l=1}^{N}{\alpha_l \sigma(w_l^T x +b_l)}\bigg\|_{m,\mathrm{p},\Omega}<\varepsilon'
\end{align}
\begin{proof}[Proof of Theorem 1]
	Let's define the function
	\begin{align}
		\label{eq:def_h}
		&h(x)=\frac{f(x)}{g(x)}
	\end{align}
	Singe $h\in \mathcal{C}^m(\Omega)$ and $\Omega\subset\mathbb{R}^n$ is compact, by the universal approximation theorem in Sobolev spaces, for any $\varepsilon>0$, there exist $N\in\mathbb{N}$, $w_l\in\mathbb{R}^n$, $b_l\in\mathbb{R}$, and $\alpha_l\in\mathbb{R}$ such that
	\begin{align}
		\label{eq:approx_h}
		&\|h(x)-\Phi(x)\|_{m,\mathrm{p},\Omega}<\varepsilon,
	\end{align}
	where
	\begin{align}
		\label{eq:def_phi}
		&\Phi(x)=\sum_{l=1}^{N}\alpha_l \sigma(w_l^T x +b_l).
	\end{align}
	
	We now estimate the Sobolev norm of the error:
	\begin{align}
		\label{eq:error_start}
		&f(x)-g(x)\Phi(x)=g(x)\big(h(x)-\Phi(x)\big).
	\end{align}
	
	Let $\alpha$ be a multi-index with $|\alpha|\le m$. By Leibniz's rule,
	\begin{align}
		\label{eq:leibniz}
		&D^\alpha\big(g(h-\Phi)\big)
		=
		\sum_{\beta\le \alpha}
		\binom{\alpha}{\beta}
		D^\beta g \, D^{\alpha-\beta}(h-\Phi).
	\end{align}
	
	Taking $L^\mathrm{p}$-norms and applying the generalized H\"older's inequality \footnote{See Exercise 6 of Chapter 8 of \cite{Wheeden}},
	\begin{align}
		\label{eq:holder0}
		&\|D^\beta g D^{\alpha-\beta}(h-\Phi)\|_{L^\mathrm{p}(\Omega)}
		\le \|D^\beta g\|_{L^\infty(\Omega)}
		\|D^{\alpha-\beta}(h-\Phi)\|_{L^\mathrm{p}(\Omega)}.
	\end{align}
	Consequently, this leads to
	\begin{align}
		\label{eq:holder}
		&\|D^\alpha(g(h-\Phi))\|_{L^\mathrm{p}(\Omega)}
		\le
		\sum_{\beta\le \alpha}
		\binom{\alpha}{\beta}
		\|D^\beta g\|_{L^\infty(\Omega)}
		\|D^{\alpha-\beta}(h-\Phi)\|_{L^\mathrm{p}(\Omega)}.
	\end{align}
	
	Note that
	\begin{align}
		\label{eq:def_M}
		&\|g\|_{m,\infty,\Omega}=\max_{|\gamma|\le m}\|D^\gamma g\|_{L^\infty(\Omega)}.
	\end{align}
	
	Then,
	\begin{align}
		\label{eq:bound_M}
		&\|D^\alpha(g(h-\Phi))\|_{L^\mathrm{p}(\Omega)}
		\le
		\|g\|_{m,\infty,\Omega}
		\sum_{\gamma\le \alpha}
		\binom{\alpha}{\gamma}
		\|D^\gamma(h-\Phi)\|_{L^\mathrm{p}(\Omega)}.
	\end{align}
	Taking into account that
	\begin{align}
		\label{eq:max_norm}
		&\|D^\gamma(h-\Phi)\|_{L^\mathrm{p}(\Omega)}\le \max_{\gamma\le \alpha}\|D^\gamma(h-\Phi)\|_{L^\mathrm{p}(\Omega)}
	\end{align}
	we obtain
		\begin{align}
		\label{eq:max_norm2}
		&\|D^\alpha(g(h-\Phi))\|_{L^\mathrm{p}(\Omega)}
		\le
		\|g\|_{m,\infty,\Omega} \max_{\gamma\le \alpha}\|D^\gamma(h-\Phi)\|_{L^\mathrm{p}(\Omega)}
		\sum_{\gamma\le \alpha}
		\binom{\alpha}{\gamma}.
	\end{align}
	Expanding the multi-index notation, we obtain
	\begin{align}
		\label{eq:binomial_sum}
		&\sum_{\gamma\le\alpha} \binom{\alpha}{\gamma}
		=
		\sum_{\gamma_1=0}^{\alpha_1}\cdots\sum_{\gamma_n=0}^{\alpha_n}
		\prod_{i=1}^n \binom{\alpha_i}{\gamma_i}
		=
		2^{|\alpha|},
	\end{align}
	where we have taken into account independence of the summation indices. Consequently, we obtain
	\begin{align}
		\label{eq:max_estimate}
		&\|D^\alpha(g(h-\Phi))\|_{L^\mathrm{p}(\Omega)}
		\le
		\|g\|_{m,\infty,\Omega}\,2^{|\alpha|}
		\max_{\gamma\le \alpha}\|D^\gamma(h-\Phi)\|_{L^\mathrm{p}(\Omega)}.
	\end{align}
	
	Since $|\alpha|\le m$, it follows that
	\begin{align}
		\label{eq:uniform_bound}
		&\|D^\alpha(g(h-\Phi))\|_{L^\mathrm{p}(\Omega)}
		\le
		\|g\|_{m,\infty,\Omega}\,2^{m}
		\max_{|\gamma|\le m}\|D^\gamma(h-\Phi)\|_{L^\mathrm{p}(\Omega)}.
	\end{align}
	
	Using the inequality
	\begin{align}
		\label{eq:max_sobolev}
		&\max_{|\gamma|\le m}\|D^\gamma u\|_{L^\mathrm{p}(\Omega)}
		\le
		\left(
		\sum_{|\alpha|\le m}
		\|D^\alpha u\|_{L^\mathrm{p}(\Omega)}^{\mathrm{p}}
		\right)^{1/\mathrm{p}}=\|u\|_{m,\mathrm{p},\Omega},
	\end{align}
	we obtain
	\begin{align}
		\label{eq:final_derivative}
		&\|D^\alpha(g(h-\Phi))\|^\mathrm{p}_{L^\mathrm{p}(\Omega)}
		\le
		\|g\|_{m,\infty,\Omega}^\mathrm{p}\,2^{m\mathrm{p}}\,\|h-\Phi\|^\mathrm{p}_{m,\mathrm{p},\Omega}.
	\end{align}
	
	where we take both sides to the $\mathrm{p}$-th power. Summing over all multi-indices with $|\alpha|\le m$, and using that \footnote{See Theorem 3 of section 3 of \cite{Haddad}}
	\begin{align}
		\label{eq:number_multiindices}
		&\sum_{|\alpha|\leq m}=\sum_{\alpha_1=0}^m
		\sum_{\alpha_2=0}^{m-\alpha_1}
		\sum_{\alpha_3=0}^{m-\alpha_1-\alpha_2}
		\cdots
		\sum_{\alpha_n=0}^{m-\alpha_1-\cdots-\alpha_{n-1}}
			=\binom{n+m}{m},
	\end{align}
	we obtain
	\begin{align}
		\label{eq:sum_estimate}
		&\sum_{|\alpha|\le m}\|D^\alpha(g(h-\Phi))\|^\mathrm{p}_{L^\mathrm{p}(\Omega)}
		\le
		\binom{n+m}{m}
		\,2^{m\mathrm{p}}\,\|g\|_{m,\infty,\Omega}^\mathrm{p}\|h-\Phi\|^\mathrm{p}_{m,\mathrm{p},\Omega}.
	\end{align}
	
	If we take both sides to the $1/\mathrm{p}$-th power, it follows that
	\begin{align}
		\label{eq:final_bound}
		&\|g(h-\Phi)\|_{m,\mathrm{p},\Omega}
		\le
		C\,\|g\|_{m,\infty,\Omega}\|h-\Phi\|_{m,\mathrm{p},\Omega},
	\end{align}
	where
	\begin{align}
		\label{eq:constant}
		&C=\binom{n+m}{m}^\frac{1}{\mathrm{p}}2^{m}.
	\end{align}
	
	Using \eqref{eq:error_start} and \eqref{eq:approx_h}, we obtain
	\begin{align}
		\label{eq:final_eps}
		&\|f-g\Phi\|_{m,\mathrm{p},\Omega}
		\le
		\varepsilon C\,\|g\|_{m,\infty,\Omega}.
	\end{align}
	
	Finally, choosing $\varepsilon'=\varepsilon C\|g\|_{m,\infty,\Omega}$ yields \eqref{eq6}.
\end{proof}

\textbf{Remark 1.} \emph{Note that \eqref{eq:final_bound} is a H\"older-type product estimate in Sobolev spaces}\\

In other words, the product of a function $g\in\mathcal{C}^m(\Omega)$ and a single-hidden-layer neural network is dense in the set 
$\{\,f\in\mathcal{C}^m(\Omega) : f/g \in \mathcal{C}^{m}(\Omega)\,\}$, 
where $\Omega \subset \mathbb{R}^{n}$ is a compact set.\\

Let us particularize Theorem 1 to the case of variational problems with boundary conditions, imposing that the function $f$, and therefore $g$, vanish at the boundary $\partial\Omega$.\\

\textbf{Corollary 1.} \emph{Let $m\in\mathbb{N}$ and $1\le p<\infty$. Let $\sigma\in\mathcal{C}^m(\mathbb{R})$ be a non-constant and bounded function. Let $\Omega\subset\mathbb{R}^n$ be compact, and let $f,g\in \mathcal{C}^m(\Omega)$ be such that}
	\begin{align}
		\label{eq:boundary_zero}
		f(x)=0,\quad g(x)=0 \quad \forall x\in\partial\Omega,
	\end{align}
	and assume that
	\begin{align}
		\label{eq:ratio_condition}
		\frac{f}{g}\in \mathcal{C}^m(\Omega).
	\end{align}
	\emph{Then, for any $\varepsilon'>0$, there exist $N\in\mathbb{N}$, $w_l\in\mathbb{R}^n$, $b_l\in\mathbb{R}$, and $\alpha_l\in\mathbb{R}$, $l=1,\ldots,N$, such that}
	\begin{align}
		\label{eq:corollary_result}
		\bigg\|f(x)-g(x)\sum_{l=1}^{N}{\alpha_l \sigma(w_l^T x +b_l)}\bigg\|_{m,\mathrm{p},\Omega}<\varepsilon'.
\end{align}\\

This corollary requires that, if $f(x)=0$ for all $x\in\partial\Omega$, then it is also required that $g(x)=0$ for all $x\in\partial\Omega$, and furthermore, $f$ must vanish at least at the same rate as $g$ in $\partial\Omega$ in order to satisfy Theorem 1.
   
To enforce the boundary condition $y(x)=b(x)$ for all $x\in\partial\Omega$, with $b\in \mathcal{C}^{1}(\Omega)$, Corollary~1 suggests the following approximation:
\begin{align}
	\label{eq14}
	&y(x)\approx B(x)+g(x)\sum_{l=1}^{N}\alpha_l\,\sigma(w_l^{T}x+b_l),
\end{align}
where $g(x)=0$ and $B(x)=b(x)$ for all $x\in\partial\Omega$. For simplicity, we choose $g(x)$ to be a polynomial $p(x)$, since polynomial functions can be easily constructed so that $p(x)=0$ for all $x\in\partial\Omega$, although other choices are possible. Note that, since $b\in\mathcal{C}^{1}(\Omega)$ and $g(x)=p(x)$ is a polynomial, choosing $B\in \mathcal{C}^{1}(\Omega)$  ensures that the proposed ansatz is an admissible function for the problem given in~\eqref{eq2}.
	
Although all theorems and corollaries are stated for $\Omega$ being a generic compact set, we assume that $\Omega$ is (piecewise) diffeomorphic to $[0,1]^{n}$ or to the unit ball $B_1^n\subset\mathbb{R}^n$ (i.e., there exists a piecewise smooth bijection between $\Omega\subset\mathbb{R}^{n}$ and either $[0,1]^{n}$ or $B_1^n$). This assumption is motivated by the fact that common differentiable activation functions exhibit significant variation when their argument lies in $[0,1]$, while they tend to saturate outside this interval, as occurs with the sigmoid or the hyperbolic tangent. Moreover, several strategies are available to construct such a diffeomorphism (see Appendix~C in \cite{Evansbook}).

In the remainder of this work, and for simplicity, we assume that $\Omega$ is (piecewise) diffeomorphic to $[0,1]^{n}$. This assumption facilitates the choice 
of a polynomial $p(x)$ satisfying $p(x)=0$ for all $x\in\partial[0,1]^{n}$, such as
\begin{align}
	\label{eq18}
	p(x) = \prod_{j=1}^{n} x_j (1 - x_j).
\end{align}
Note that this choice of polynomial ensures that $p(x)\to 0$ linearly as $x_j\to 0$ or $x_j\to 1$ (i.e., $p(x)$ vanishes linearly as $x$ approaches the 
$j$-th face of the hypercube). Since $p(x)$ vanishes linearly near $\partial[0,1]^{n}$, the solution of the 
variational problem $y(x)$ must verify $(y(x)-b(x))\rightarrow 0$ at least at the same rate as $p(x)$ when approaching the boundary, as required to satisfy Theorem~1. This last condition is fulfilled, since $y(x)$ and $b(x)$ belong to $\mathcal{C}^1(\Omega)$ and therefore satisfy Taylor theorem.

The case of $B_1^n$ could be easily handled choosing $g(x)=1-\|x\|^2_2$, but this case will not be considered in this work.

\section{Ansatz Construction and Variational Minimization Procedure}\label{sec3}

\hspace{2em} Let us consider the differentiable transformation $T:[0,1]^n\rightarrow\Omega$ with $\Omega\subset\mathbb{R}^n$ such that if $u\in\partial[0,1]^{n}$ then $T(u)=x\in\partial\Omega$, where $x=(x_1,\ldots,x_n)$ and $u=(u_1,\ldots,u_n)$, such that $x_j=T_j(u_1,\ldots,u_n)$ with $T_j:[0,1]^n\rightarrow\mathbb{R}$. Applying this transformation to \eqref{eq2} leads to the following result:
\begin{align}
	\label{eq15}
	\hat{S}=\int_{[0,1]^n}{\mathcal{L}\left(\hat{y}(u),(J_T(u)^{-1})^T\frac{\partial \hat{y}(u)}{\partial u},T(u)\right)|\mathrm{det}(J_{T}(u))|du}
\end{align}
where $\left(J_{T}(u)\right)_{ij}=\frac{\partial T_i}{\partial u_j}$. In \eqref{eq15}, the expressions $y(x)=\hat{y}(u)$, $\frac{\partial y(x)}{\partial x}=(J_T(u)^{-1})^T\frac{\partial \hat{y}(u)}{\partial u}$ and $x=T(u)$ have been taken into account. For $y(x)\in \mathcal{C}^1$ to be an admissible function, it must satisfy $y(x)=b(x)$ $\forall x\in\partial\Omega$. Therefore, the $T$ differentiable transformation has to ensure $\hat{y}(u)=b(T(u))$ $\forall u\in\partial[0,1]^{n}$. Let us denote $\hat{b}(u)=b(T(u))$. According to the previous section, we propose the following ansatz:
\begin{align}
	\label{eq16}
	&\hat{y}(u,\theta)= B(u)+p(u)N_{\mathrm{net}}(u,\theta)
\end{align}
where $\theta=(\alpha_{1},\ldots,\alpha_{N},w_{1},\ldots,w_{N},b_{1},\ldots,b_{N})$, and $N_{\mathrm{net}}(u,\theta)$ represents single-hidden-layer neural networks with $N$ neurons:
\begin{align}
	\label{eq17}
	&N_{\mathrm{net}}(u,\theta)= \sum_{l=1}^{N}{\alpha_{l} \tanh\left(w_{l}^T u +b_{l}\right)}
\end{align}
where $N\in\mathbb{N}$, $w_{l}\in\mathbb{R}^n$, $b_{l}\in \mathbb{R}$, $\alpha_{l}\in\mathbb{R}$, the polynomial $p(u)$ is defined as \eqref{eq18} and where the term $B(u)$ satisfies $B(u)=\hat{b}(u)$ $\forall u\in\partial[0,1]^n$. This can be achieved through transfinite interpolation \cite{Gordon},\cite{Varady}, following the inclusion-exclusion principle \cite{Hall}. For instance, in the one-dimensional case, given $\hat{b}(0)$ and $\hat{b}(1)$, this leads to: 
\begin{align}
	\label{eq19}
	&B(u)=(1-u)\hat{b}(0)+u\hat{b}(1)
\end{align}
and for the bidimensional case (also known as the Coons patch \cite{Coons}), given $\hat{b}(0,u_2)$, $\hat{b}(1,u_2)$, $\hat{b}(u_1,0)$, $\hat{b}(u_1,1)$, $\hat{b}(0,0)$, $\hat{b}(0,1)$, $\hat{b}(1,0)$, and $\hat{b}(1,1)$, this leads to:
\begin{align}
	\nonumber
	&B(u)=\\
	\nonumber
	&\left[(1-u_1)\hat{b}(0,u_2)+u_1\hat{b}(1,u_2)+(1-u_2)\hat{b}(u_1,0)+u_2\hat{b}(u_1,1)\right]-\\
	\label{eq20}
	&\left[(1-u_1)(1-u_2)\hat{b}(0,0)+u_1(1-u_2)\hat{b}(1,0)+\right.\left.(1-u_1)u_2\hat{b}(0,1)+u_1u_2\hat{b}(1,1)\right]
\end{align}


In what follows, for simplicity, we will consider a convex Lagrangian with a convex domain. Under this consideration, the condition of null variation of the action (i.e., $\delta \hat{S}=0$) is equivalent to minimizing the action $\hat{S}$ \cite{Dacorogna},\cite{Rindler}. Therefore, we will aim to solve the following problem:
\begin{align}
	\label{eq21}
	&\min_{\hat{y}(u)}\left\{\hat{S}(\hat{y}(u)) : \hat{y}(u)=\hat{b}(u),\ \forall u\in\partial[0,1]^{n}\right\}
\end{align}

If we use the proposed ansatz given by \eqref{eq16}, the equivalent problem given by \eqref{eq21} becomes a minimization problem with respect to the parameters of the neural network $\theta$. The advantage of this approach over the Deep Ritz method is that penalization is not necessary for the boundary condition, as it is inherently satisfied by the proposed ansatz (i.e., the proposed ansatz given by \eqref{eq16} constitutes a set of admissible functions).

The minimization of the action will be carried out using the gradient descent method, where the parameters $\theta$ are updated iteratively as follows:
\begin{align}
	\label{eq22}
	&\theta^{k+1}=\theta^{k}-\eta \frac{\partial\hat{S}(\theta^k)}{\partial \theta}
\end{align}
where $\theta^k=(\theta_1^k,\ldots,\theta_{3N}^k)$ contains the neural network parameters $\alpha_{1},\ldots,\alpha_{N},w_{1},\ldots,w_{N},b_{1},\ldots,b_{N}$ for the $k$th iteration, $\eta\in\mathbb{R}$ is a learning rate (along results section, $\eta=0.001$ will be used), and $\frac{\partial\hat{S}}{\partial \theta}=(\frac{\partial\hat{S}}{\partial \theta_1},\ldots,\frac{\partial\hat{S}}{\partial \theta_{3N}})$ where
\begin{align}
	\label{eq23}
	\frac{\partial\hat{S}}{\partial \theta}=\int_{[0,1]^{n}}{\left[\frac{\partial\mathcal{L}}{\partial \hat{y}}\cdot\frac{\partial\hat{y}}{\partial \theta}+\frac{\partial\mathcal{L}}{\partial \left(\frac{\partial\hat{y}}{\partial u}\right)}\cdot\frac{\partial^2\hat{y}}{\partial \theta \partial u}\right]|\mathrm{det}(J_{T}(u))|du}
\end{align}
where $\left(\frac{\partial\hat{y}}{\partial \theta}\right)_{s}=\frac{\partial\hat{y}}{\partial \theta_s}$, $\left(\frac{\partial\mathcal{L}}{\partial \left(\frac{\partial\hat{y}}{\partial u}\right)}\right)_{j}=\frac{\partial\mathcal{L}}{\partial \left(\frac{\partial\hat{y}}{\partial u_j}\right)}$, $\left(\frac{\partial^2\hat{y}}{\partial \theta \partial u}\right)_{sj}=\frac{\partial^2\hat{y}}{\partial \theta_s \partial u_j}$. The derivatives of $\hat{y}$ are calculated from the proposed ansatz given by \eqref{eq16} as follows:
\begin{align}
	\label{eq24}
	&\frac{\partial \hat{y}}{\partial \theta_s}=p(u)\frac{\partial N_{\mathrm{net}}}{\partial\theta_s}\\
	\label{eq25}
	&\frac{\partial^2\hat{y}}{\partial \theta_s \partial u_j}=\frac{\partial p}{\partial u_j}\frac{\partial N_{\mathrm{net}}}{\partial \theta_s}+p(u)\frac{\partial^2 N_{\mathrm{net}}}{\partial \theta_s\partial u_j}
\end{align}
where 
\begin{align}
	\label{eq26}
	&\frac{\partial p}{\partial u_j}=(1-2u_j)\prod_{i\neq j}^n u_i(1-u_i)
\end{align}

The computation of the integral given by \eqref{eq15} and \eqref{eq23} is approximated using conventional quadrature rules, such as Gauss-Legendre, or specialized quadrature rules for cases involving singular integrands. 

\section{Results}\label{sec4}
\hspace{2em} In this section, we present some examples where the results obtained using the proposed method are compared with those achieved by the Deep Ritz method, as shown in \cite{DRM}. All results were obtained using custom-developed code written in Python and PyTorch \cite{Paszke}, leveraging the efficient neural network library to access the required evaluation functions and derivatives. 

\subsection{Example 1}\label{sub1sec4}
\hspace{2em} In this section, we demonstrate the application of the proposed method for a simple action, in order to illustrate didactically how to proceed.
In this example, we aim to solve the following problem:
\begin{align}
	\label{eq27}
	\min_{y(x)}\left\{ S=\int_{a}^{b}{\left[y(x)+\left(\frac{dy}{dx}\right)^2\right]dx} : y(a)=A,\ y(b)=B\right\} 
\end{align}
where $x\in[a,b]$ and $y(x)\in\mathbb{R}$. Let us consider the linear transformation $T:[0,1]\rightarrow [a,b]$ defined as $x=(b-a)u+a$, where $u\in[0,1]$. Applying this linear transformation to the problem defined in \eqref{eq27} leads to:
 \begin{align}
 	\label{eq28}
 	\min_{\hat{y}(u)}\left\{ \hat{S}=\int_{0}^{1}{\left[(b-a)\hat{y}(u)+\frac{1}{b-a}\left(\frac{d\hat{y}}{du}\right)^2\right]du} : \hat{y}(0)=A,\ \hat{y}(1)=B\right\} 
 \end{align}
\indent In order to minimize the action with a set of admissible functions, we propose the following ansatz, in accordance with \eqref{eq16}, and following \eqref{eq19}:
\begin{align}
	\label{eq29}
	&\hat{y}(u,\theta)=A(1-u)+Bu+u(1-u)N_{\mathrm{net}}(u,\theta)
\end{align}
where $N_{\mathrm{net}}(u,\theta)$ is defined as in \eqref{eq17}. Note that $\hat{y}(0,\theta)=A$ and $\hat{y}(1,\theta)=B$, using the proposed ansatz in \eqref{eq29} (i.e., the set of $\hat{y}(u,\theta)\in \mathcal{C}^1$ provided by \eqref{eq29} constitutes a set of admissible functions). According to \cite{DRM}, the minimization problem via the Deep Ritz method is reduced to minimizing the action given by:
 \begin{align}
	\nonumber
	&\hat{S}_{\mathrm{DR}}=\int_{0}^{1}{\left[(b-a)N_{\mathrm{net}}(u,\theta)+\frac{1}{b-a}\left(\frac{\partial N_{\mathrm{net}}(u,\theta)}{\partial u}\right)^2\right]du}\\
	\label{eq30}
	&+\beta\left[\left(N_{\mathrm{net}}(0,\theta)-A\right)^2+\left(N_{\mathrm{net}}(1,\theta)-B\right)^2\right]
\end{align}

In the following, we will consider $a=A=0$ and $b=B=10$. Since the Lagrangian is convex with a convex domain, the exact solution can be obtained by solving the Euler-Lagrange equations. This exact solution, for these values of $a$, $b$, $A$, and $B$, is given by:
\begin{align}
	\label{eq31}
	\hat{y}_{\mathrm{exact}}(u)=25u^2-15u
\end{align}

\begin{figure}[H]
	\centering
	\includegraphics [width=8 cm]{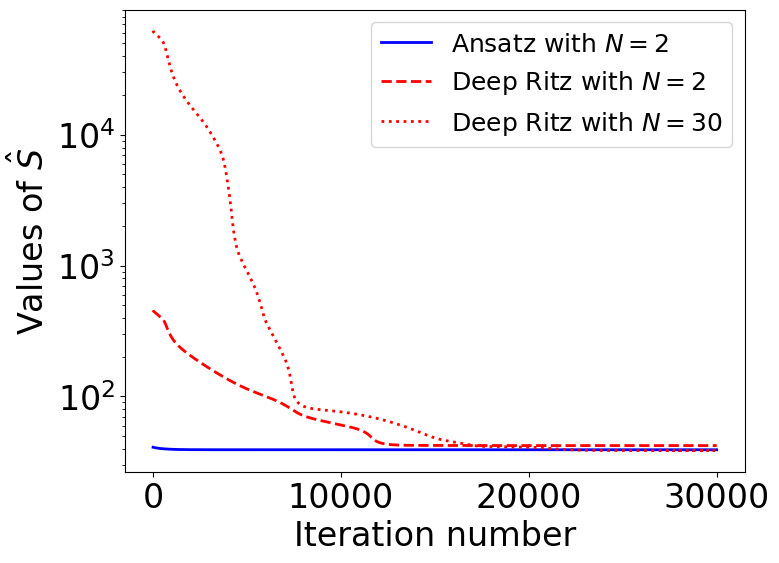}
	\caption{\label{fig1} Values of $\hat{S}$ with respect to the iteration number of the gradient descent method. These values are presented for three cases: case $N=2$, where $\hat{S}$ is minimized using the ansatz given by \eqref{eq29}; case $N=2$, where $\hat{S}_{\mathrm{DR}}$ is minimized; and case $N=30$, where $\hat{S}_{\mathrm{DR}}$ is minimized.}
\end{figure}

Figure \ref{fig1} shows the values of $\hat{S}$ with respect to the iteration number of the gradient descent method. Three cases have been considered: case $N=2$ when $\hat{S}$ is minimized using the ansatz given by \eqref{eq29}; case $N=2$ when $\hat{S}_{\mathrm{DR}}$ is minimized; and case $N=30$ when $\hat{S}_{\mathrm{DR}}$ is minimized. The value of $\beta=50$ has been used for the minimization of $\hat{S}_{\mathrm{DR}}$. With this value of $\beta$, the boundary values of the $N_{\mathrm{net}}$ reached in the last iteration are $N_{\mathrm{net}}(0,\theta)=-0.04$ and $N_{\mathrm{net}}(1,\theta)=9.94$ for $N=2$, and $N_{\mathrm{net}}(0,\theta)=-0.03$ and $N_{\mathrm{net}}(1,\theta)=9.93$ for $N=30$. The initial values of the $\theta$ parameters were randomly sampled from a Gaussian distribution with a mean of 10 and a standard deviation of 0.1, and these initial values were identical for all cases considered.
The obtained results in Figure \ref{fig1} show that faster convergence is achieved when $\hat{S}$, using the ansatz given in \eqref{eq29}, is minimized. Specifically, four significant digits are reached at iteration 4200. In comparison, using the Deep Ritz method, four significant digits are reached at iterations 29200 and 29400 for $N=2$ and $N=30$, respectively.

It is worth noting that the values of $\hat{S}$ obtained with the Deep Ritz method for $N=30$ are the lowest at the final iteration. This might suggest that the solution closest to the exact solution is provided by the Deep Ritz method with $N=30$. However, as shown in subsequent figures, the solution closest to the exact solution is actually provided by the proposed method.

This observation may appear counterintuitive, since the Deep Ritz method with $N=30$ achieves the lowest value of $\hat{S}$. Nevertheless, it is important to consider that the boundary values obtained using the Deep Ritz method with $N=30$ are close to the exact values ($N_{\mathrm{net}}(0,\theta)=-0.03$ and $N_{\mathrm{net}}(1,\theta)=9.93$), but not exact. In fact, with these boundary values, the lowest values of $\hat{S}$ achieved by the Deep Ritz method with $N=30$ are very close to the values of $\hat{S}$ obtained by the proposed method at the final iteration. This indicates that a failure to exactly satisfy the boundary conditions can lead to this type of misleading result.

\begin{figure}[H]
	\centering
	\begin{tabular}{c}
		\includegraphics[width=6.0 cm]{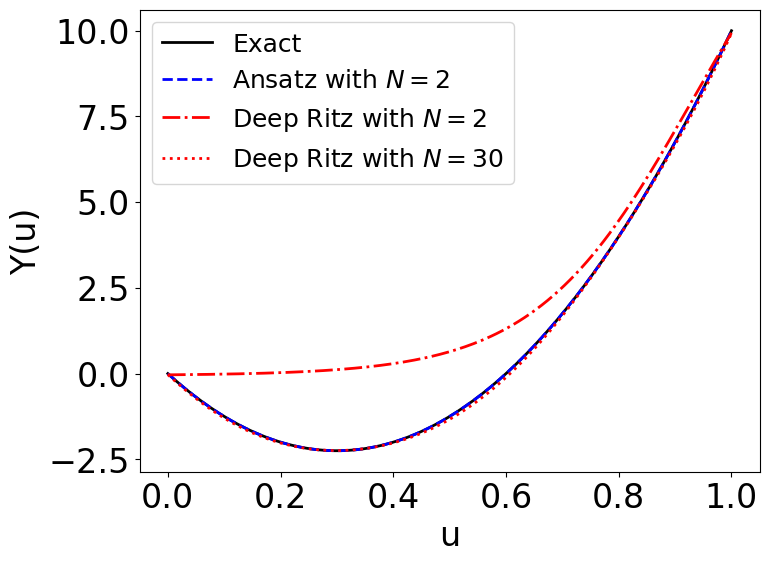} \includegraphics[width=6.0 cm]{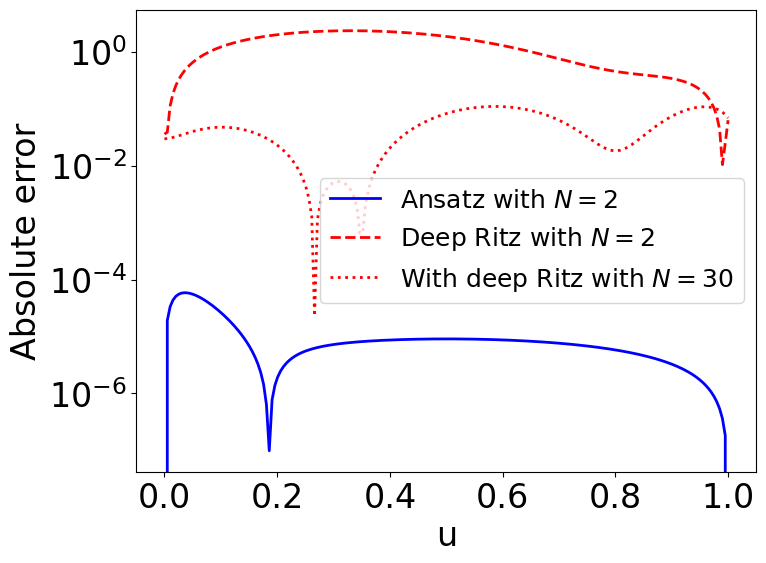}\\
		\hspace{1cm} (a) \hspace{5.5cm} (b)\\
	\end{tabular}
	\caption{\label{fig2} (\textbf{a}) $\hat{y}(u)$ values for the $\theta$ values obtained in the final iteration; (\textbf{b}) absolute error with respect to the exact solution given by \eqref{eq31}. Three cases are considered: the proposed method using the ansatz provided in \eqref{eq29}, the Deep Ritz method with $N=2$, and the Deep Ritz method with $N=30$.}
\end{figure}

Figure \ref{fig2}(a) illustrates the $\hat{y}(u)$ values with respect to $u\in[0,1]$ for the $\theta$ values obtained in the final iteration. Figure \ref{fig2}(b) displays the absolute error relative to the exact solution given by \eqref{eq31}. The three cases are once again considered: the case of the proposed method using the ansatz given in \eqref{eq29}, the case of the Deep Ritz method with $N=2$, and the case of the Deep Ritz method with $N=30$.
Although the boundary condition is approximately satisfied by the solution obtained using the Deep Ritz method for $N=2$, achieving $N=30$ is necessary to approximate the solution to the exact solution (i.e., the differences between the Deep Ritz solution and the exact solution are unacceptable when $N=2$). In fact, the absolute error achieved by the Deep Ritz method is approximately 0.01 when $N=30$, whereas the error obtained by the proposed method is less than 0.0001 when $N=2$. These results validate the use of the ansatz, as the absolute errors of the solution obtained are 100 times smaller than those of the Deep Ritz method, despite employing a very simple neural network with a single hidden layer and $N=2$.

\subsection{Example 2}\label{sub2sec4}

\hspace{2em} In this section, we aim to solve the following problem:
\begin{align}
	\label{eq32}
	&\min_{y(x_1,x_2)}\left\{\begin{matrix}
		S=\int_{\Omega}{\left[||\nabla y(x_1,x_2)||^2-f(x_1,x_2)y(x_1,x_2)\right]dx}\\
		\\
		y(x_1,x_2)=b(x_1,x_2), \forall (x_1,x_2)\in\partial\Omega
	\end{matrix} \right\} 
\end{align}
where $\Omega=[0,1]\times[0,1]$, the norm $||\cdot||$ is the Euclidean norm, and $f(x_1,x_2)$ is defined as follows:
\begin{align}
	\label{eq33}
	&f(x_1,x_2)=e^{-x_1}(2-x_1-x_2^3-6x_2)
\end{align}
and where:
\begin{align}
	\label{eq34}
	&b(x_1,x_2)=\left\{\begin{matrix}
		\hspace{0cm} x_2^3 \hspace{1.7cm} \mathrm{if} \hspace{0.3cm} x_1=0\\
		e^{-1}(1+x_2^3) \hspace{0.3cm} \mathrm{if} \hspace{0.3cm} x_1=1\\
		\hspace{0cm} x_1e^{-x_1} \hspace{1.1cm} \mathrm{if} \hspace{0.3cm} x_2=0\\
		(x_1+1)e^{-x_1} \hspace{0.3cm} \mathrm{if} \hspace{0.3cm} x_2=1\\
		0 \hspace{2cm} \mathrm{otherwise}
	\end{matrix} \right.\
\end{align}
The exact solution is provided as follows:
\begin{align}
	\label{eq35}
	&y_{\mathrm{exact}}(x_1,x_2)=e^{-x_1}\left(x_1+x_2^3\right)
\end{align}
In order to minimize the action with a set of admissible functions, we propose the following ansatz, in accordance with \eqref{eq16}:
\begin{align}
	\label{eq36}
	&y(x_1,x_2,\theta)=B(x_1,x_2)+x_1(1-x_1)x_2(1-x_2)N_{\mathrm{net}}(x_1,x_2,\theta)
\end{align}
where $B(x_1,x_2)$ is directly derived from \eqref{eq20} by substituting $x_1,x_2$ in place of $u_1,u_2$. It is important to note that, in this case, we work directly with variables $x_1$ and $x_2$ to simplify the $T$ transformation described in the previous section. According to \cite{DRM}, the minimization problem in the Deep Ritz method is reduced to minimizing the action given by:
\begin{align}
	\nonumber
	&S_{\mathrm{DR}}=\int_{\Omega}{\left[||\nabla N_{\mathrm{net}}(x_1,x_2,\theta)||^2-f(x_1,x_2)N_{\mathrm{net}}(x_1,x_2,\theta)\right]dx}\\
	\label{eq37}
	&+\beta\int_{\partial\Omega}{\left[N_{\mathrm{net}}(x_1,x_2,\theta)-b(x_1,x_2)\right]^2dx}
\end{align}

\begin{figure}[H]
	\centering
	\includegraphics [width=8 cm]{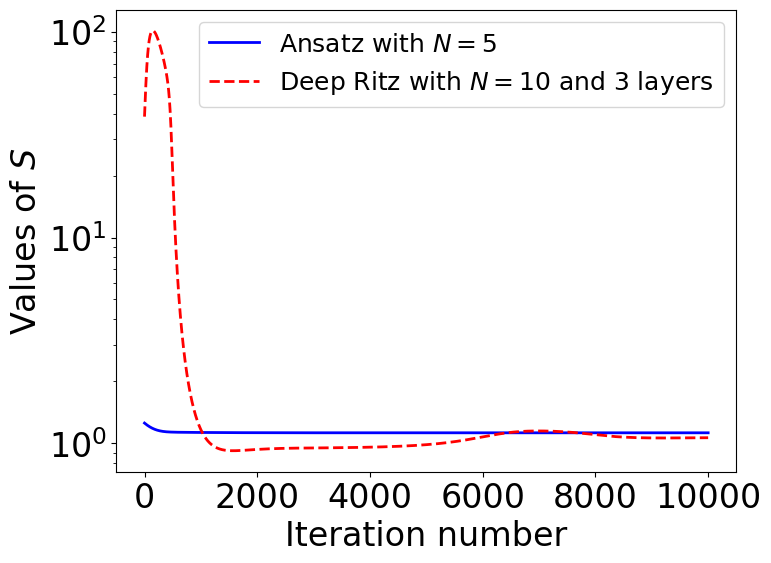}
	\caption{\label{fig3} Values of $S$ with respect to the iteration number of the gradient descent method are presented for two cases: the case of $N=10$, where $S$ is minimized using the ansatz provided in \eqref{eq36}, and the case of $N=10$ with an additional hidden layer, where $S_{\mathrm{DR}}$ is minimized.}
\end{figure}

Figure \ref{fig3} illustrates the values of $S$ with respect to the iteration number of the gradient descent method. Two cases have been considered: the case of $N=5$, where $S$ is minimized using the ansatz given by \eqref{eq36}, and the case of $N=10$, with an additional hidden layer, where $S_{\mathrm{DR}}$ is minimized. This additional hidden layer was necessary to achieve acceptable results in the minimization of $S_{\mathrm{DR}}$. A value of $\beta=30$ was used during the minimization of $S_{\mathrm{DR}}$. The initial values of the $\theta$ parameters were randomly sampled from a Gaussian distribution with a mean of 1 and a standard deviation of 0.1, with these initial values being identical across all considered cases.
The obtained results in Figure \ref{fig3} show that faster convergence is achieved when $S$, using the ansatz given in \eqref{eq36}, is minimized. Specifically, four significant digits are reached at iteration 2800, whereas with the Deep Ritz method, four significant digits are reached at iteration 9800.

It is worth noting that the lowest value of $S$ is achieved by the Deep Ritz method at approximately 2000 iterations. However, this is accomplished at the expense of deviating the solution's boundary values from those required by the boundary condition in the optimization process. This demonstrates another advantage of using an ansatz that exactly satisfies the boundary conditions, as it prevents such misleading outcomes.

We would like to point out that, as observed in the figure \ref{fig3}, the value of $S$ obtained with the Deep Ritz method is not monotonically decreasing with respect to the iteration number, whereas the value of $S$ produced by the proposed method is. This behavior is expected, since the Deep Ritz method does not minimize $S$ directly but rather the functional $S_{\mathrm{DR}}$ defined in~\eqref{eq37}, which includes the penalization term enforcing the boundary conditions. In contrast, the functional $S_{\mathrm{DR}}$ itself is monotonically decreasing along the iterations, as expected from the application of gradient descent. 

\begin{figure}[H]
	\centering
	\begin{tabular}{c}
		\includegraphics[width=6.0 cm]{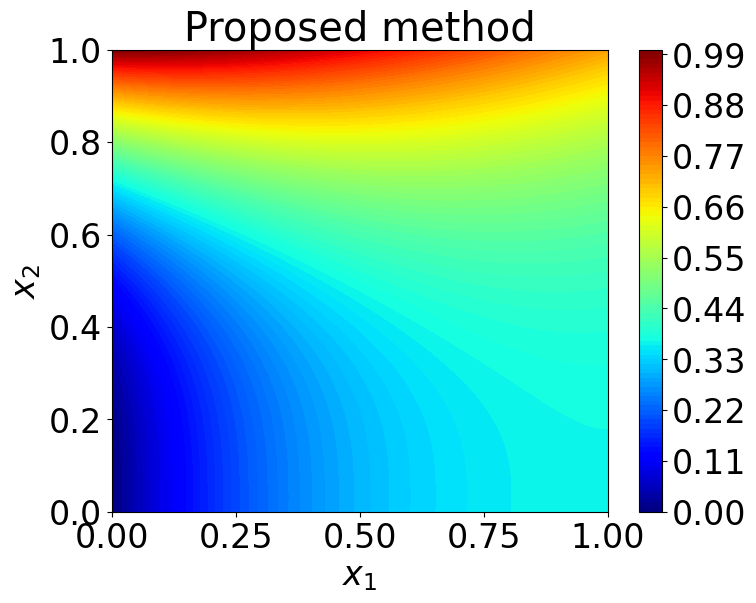} \includegraphics[width=6.0 cm]{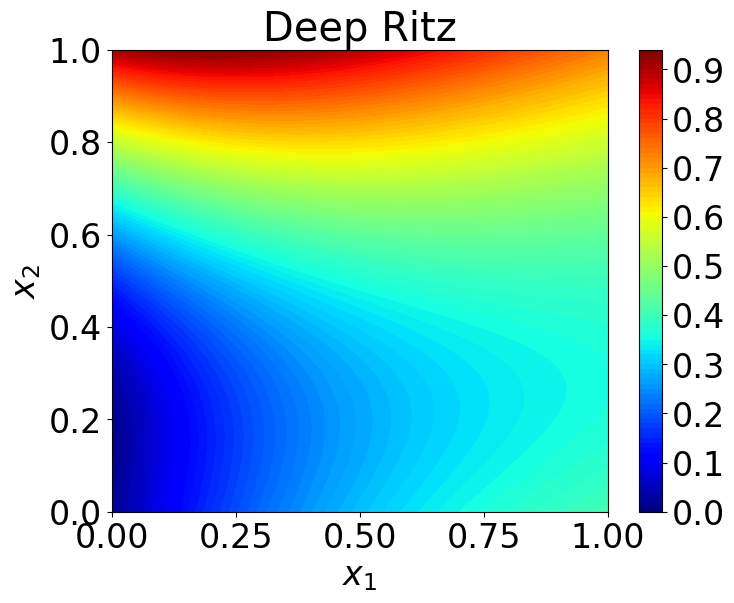}\\
		(a) \hspace{6.5cm} (b)\\
	\end{tabular}
	\caption{\label{fig4} $y(x_1,x_2,\theta)$ values corresponding to the $\theta$ values obtained in the final iteration are shown for (\textbf{a}) the proposed method and (\textbf{b}) the Deep Ritz method.}
\end{figure}

\begin{figure}[H]
	\centering
	\begin{tabular}{c}
		\includegraphics[width=6.5 cm]{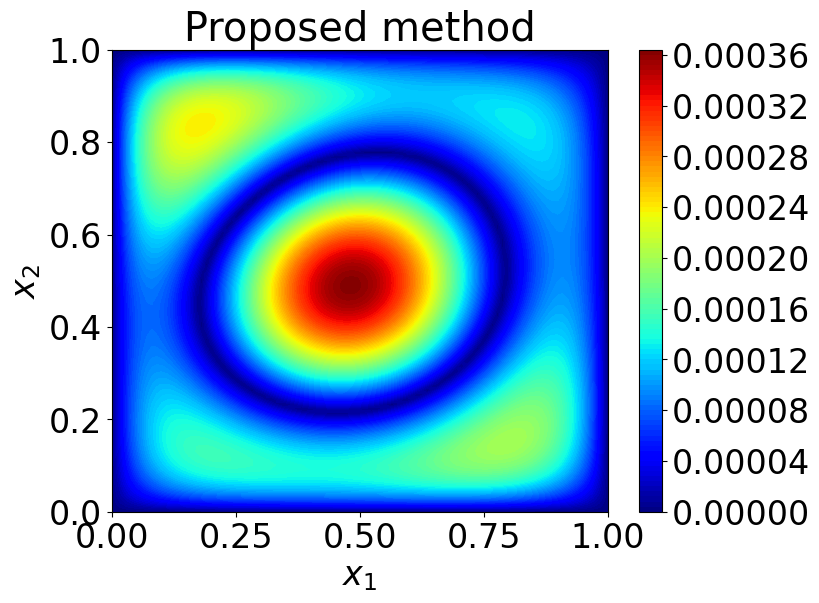} \includegraphics[width=6.0 cm]{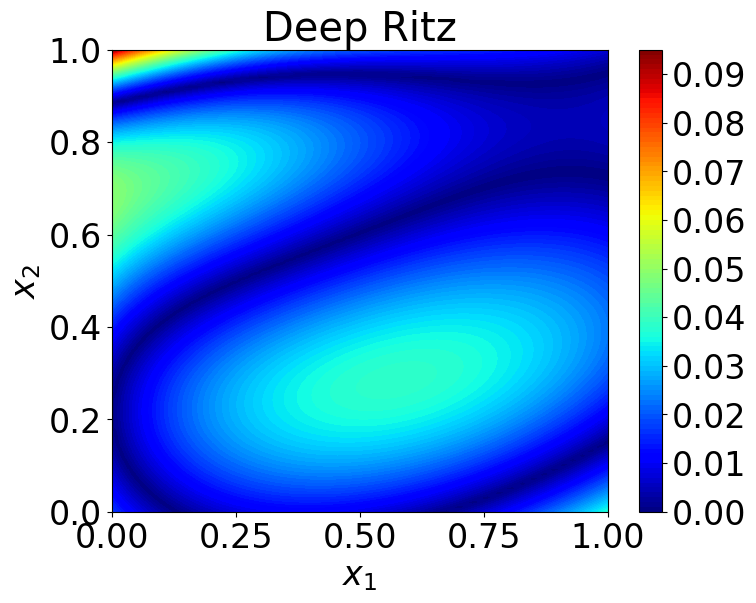}\\
		(a) \hspace{6.5cm} (b)\\
	\end{tabular}
	\caption{\label{fig5} Absolute errors relative to the exact solution given by \eqref{eq35}, obtained using (\textbf{a}) the proposed method and (\textbf{b}) the Deep Ritz method. Note that the scale of colormap is different in both figures.}
\end{figure}

Figure \ref{fig4}(a) illustrates the $y(x_1,x_2)$ values with respect to $x_1$ and $x_2$ for the $\theta$ values obtained in the final iteration of the proposed method, whereas Figure \ref{fig4}(b) shows the $y(x_1,x_2)$ values for the $\theta$ values obtained in the final iteration of the Deep Ritz method.
Although the results shown in Figures \ref{fig4}(a) and \ref{fig4}(b) are qualitatively similar, significant differences can be observed when compared to the exact solution given by \eqref{eq35}. These differences are highlighted in Figures \ref{fig5}(a) and \ref{fig5}(b), which display the absolute errors relative to the exact solution \eqref{eq35} for the proposed method and the Deep Ritz method, respectively.
The absolute errors obtained by the proposed method are less than 0.0005, whereas those obtained by the Deep Ritz method are bounded above by 0.09. Note that in the proposed method the error is zero at the boundaries, whereas in the Deep Ritz method the maximum error is attained in $(x_1,x_2)=(0, 1)$. It is also worth highlighting that the results achieved by the proposed method were obtained using a neural network with a single hidden layer and $N=5$, whereas the Deep Ritz method required two hidden layers with $N=10$ to achieve comparable results. These findings further underscore the value of employing an ansatz that precisely satisfies the boundary conditions, as it eliminates the need for additional complexity while ensuring accuracy.

\subsection{Example 3}\label{sub3sec4}
\hspace{2em} In this section, we address the task of finding the minimum eigenvalue for the quantum oscillator problem under Dirichlet boundary conditions. For this type of problem, the eigenvalue can be approximated as follows:
\begin{align}
	\label{eq38}
	&\lambda[y(x)]=\frac{\int_{\Omega}{\left[||\nabla y(x)||^2+||x||^2y(x)^2\right]dx}}{\int_{\Omega}{y(x)^2dx}}
\end{align}
where $y(x)^2$ is interpreted as the probability density of locating a particle at a given position. This interpretation leads to the normalization condition $\int_{\Omega}{y(x)^2dx}=1$. Note that expression \eqref{eq38} is the Rayleigh quotient \cite{Brezis} associated with the energy operator $\hat{E}=-\nabla^2 y(x)+||x||^2y(x)$. It is important to point out that, in quantum mechanics, the square norm is associated with the product of a function and its complex conjugate, since the wave function (in our case, $y(x)$) is generally a complex function.
Therefore, $y(x)$ is determined by solving the following problem:
\begin{align}
	\label{eq39}
	&\min_{y(x_1,x_2)}\left\{\begin{matrix}
		S=\frac{\int_{\Omega}{\left[||\nabla y(x)||^2+||x||^2y(x)^2\right]dx}}{\int_{\Omega}{y(x)^2dx}}\\
		\\
		y(x)=0, \forall x\in\partial\Omega
	\end{matrix} \right\} 
\end{align}
In the case of Deep Ritz method, the minimization problem is given by: 
\begin{align}
	\label{eq40}
	&S_{\mathrm{DR}}=\frac{\int_{\Omega}{\left[||\nabla y(x)||^2+||x||^2y(x)^2\right]dx}}{\int_{\Omega}{y(x)^2dx}}+\beta\int_{\partial\Omega}{y(x)^2dx}
\end{align}
Where $y(x)=N_{\mathrm{net}}(x,\theta)$ in \eqref{eq40}. In \cite{DRM}, an example of finding the minimum eigenvalue for a quantum oscillator was presented for $\Omega=[-3,3]^d$ by adding a penalization term $\left(\int_{\Omega}{y(x)^2dx}-1\right)^2$. However, after implementing this minimization with the added penalization, we observed that although the penalization improves the stabilization of the optimization process, the convergence remains roughly similar to the case of minimizing $S_{\mathrm{DR}}$ without the penalization term $\left(\int_{\Omega}{y(x)^2dx}-1\right)^2$:

\indent So, we compare the results obtained from the minimization of $S_{\mathrm{DR}}$ with those from the minimization of $S$, as defined by \eqref{eq37}, and the following proposed ansatz:
\begin{align}
	\label{eq41}
	&y(x)=p(x)N_{\mathrm{net}}(x,\theta)
\end{align}
where:
\begin{align}
	\label{eq42}
	&p(x)=\prod_{j=1}^d\left(1-\left(\frac{x_i}{3}\right)^2\right)
\end{align}
\indent We consider the case where $d=3$, for which the exact minimum eigenvalue is $\lambda_{\mathrm{exact}}=3$. 

\begin{figure}[H]
	\centering
	\includegraphics [width=8 cm]{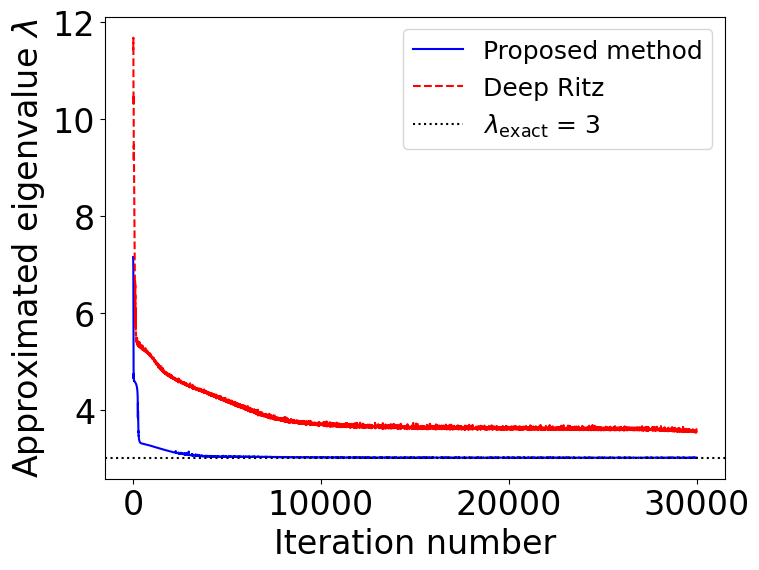}
	\caption{\label{fig6} Values of $\lambda$ via \eqref{eq38} with respect to the iteration number of the gradient descent method are shown for two cases: one where $S$ is minimized using the ansatz provided in \eqref{eq41}, and another where $\hat{S}_{DR}$ is minimized.}
\end{figure}

Figure \ref{fig6} illustrates the values of $\lambda$ via \eqref{eq38} with respect to the iteration number of the gradient descent method. Two cases have been considered: one where $S$ is minimized using the ansatz provided in \eqref{eq41}, and another where $S_{\mathrm{DR}}$ is minimized. In both cases, $N=50$ is used, with $\beta=15$. The initial values of the $\theta$ parameters were randomly sampled from a Gaussian distribution with a mean of 0 and a standard deviation of 0.1, and these initial values were identical for both considered cases.
We observe that convergence is achieved more rapidly with the proposed method than with the Deep Ritz method. Specifically, a value of $\lambda=3.0096$ (i.e., a relative error of 0.3\% with respect to the exact value) is obtained using the proposed method, whereas the Deep Ritz method reaches $\lambda=3.5939$ (i.e., a relative error of 19.8\% with respect to the exact value) in the final iteration. Moreover, spurious oscillations appear in the values of $\lambda$ for the Deep Ritz method. To further investigate this, we performed numerical simulations using the Deep Ritz method with an added penalization term $\left(\int_{\Omega}{y(x)^2dx}-1\right)^2$. While this adjustment eliminates the oscillations, the overall convergence behavior remains similar. 

\section{Conclusions}\label{sec5}

\hspace{2em} An ansatz satisfying the boundary conditions for the variational problem has been proposed. This ansatz eliminates the need for a penalty term to account for the boundary conditions, as required in the Deep Ritz method. The ansatz comprises a term representing boundary behavior and another term formed by the product of a neural network and a polynomial. The polynomial is designed to vanish at the boundaries of the domain, ensuring that the boundary conditions are satisfied. Within the rest of the domain, the behavior of the ansatz is primarily governed by the product of the neural network and the polynomial.


A central contribution of this work is that all supporting theorems and corollaries have been established in Sobolev spaces. This is particularly 
relevant because the natural notion of approximation in variational problems involves not only the function itself but also its derivatives. Accordingly, proving density and approximation properties in Sobolev spaces provides a mathematically rigorous justification for the admissibility and effectiveness of the proposed ansatz within the Ritz framework. In particular, we have shown that the product of a polynomial that vanishes linearly near the domain boundary and a single-hidden-layer neural network is dense in the corresponding Sobolev space. This result guarantees that the proposed ansatz is expressive enough to approximate the solution of the variational problem with respect to the natural Sobolev norm.

Several drawbacks have been addressed by the proposed ansatz in various examples, demonstrating its effectiveness in overcoming key challenges:

\begin{itemize}
	\item  The dependence on hyperparameters is unnecessary for enforcing the boundary conditions.
	\item  The proposed ansatz effectively reduces complexity while maintaining accuracy, highlighting its practical advantage in addressing variational problems.
	\item While the Deep Ritz method achieves the lowest action values, the proposed method offers the solution closest to the exact one, emphasizing the critical role of accurately satisfying boundary conditions to prevent misleading outcomes in the optimization of the action.
\end{itemize}

\section*{Declarations}

\begin{itemize}
\item Funding: This work was supported by the 'Plan Estatal de Investigaci\'on', under the project PID2022-138144NB-I00 entitled 'Simetr\'ia y coherencia en sistemas cr\'iticos cu\'anticos'.
\item Conflict of interest/Competing interests: Not applicable
\item Ethics approval and consent to participate: Not applicable
\item Consent for publication: Not applicable
\item Data availability: Not applicable 
\item Materials availability: Not applicable
\item Code availability: Available under a reasonable request to the authors. 
\item Author contribution: Not applicable
\end{itemize}


\end{document}